\pdfoutput=1

\documentclass[Afour,sageh,times]{sagej}

\usepackage{moreverb,url}
\usepackage{subcaption}
\usepackage{textcomp, gensymb}
\usepackage{multirow}
\usepackage{caption}
\usepackage{natbib}
\usepackage[T1]{fontenc}
\usepackage{graphicx}
\usepackage{soul}
\usepackage{pifont}
\usepackage{bbding}
\usepackage{soul,color}
\usepackage{pgfplots}
\usepackage{stfloats}

\usepackage{tabu}
\usepackage{tikz}
\usetikzlibrary{calc,decorations.markings}
\usetikzlibrary{intersections}
\usetikzlibrary{decorations.pathmorphing}
\tikzset{snake it/.style={decorate, decoration=snake}}

\newcommand\BibTeX{{\rmfamily B\kern-.05em \textsc{i\kern-.025em b}\kern-.08em
T\kern-.1667em\lower.7ex\hbox{E}\kern-.125emX}}

\def\volumeyear{2024}

\begin{document}

\runninghead{Sethuraman et al.}

\title{Machine Learning for Shipwreck Segmentation from Side Scan Sonar Imagery: Dataset and Benchmark}

\author{Advaith V. Sethuraman$^{*,1}$, Anja Sheppard$^{*,1}$, Onur Bagoren$^{1}$, Christopher Pinnow$^{2}$, Jamey Anderson$^{2}$, Timothy C. Havens$^{2}$, and Katherine A. Skinner$^{1}$}

\affiliation{This work is supported by the NOAA Ocean Exploration Program under award NA21OAR0110196.\\
\affilnum{*}denotes equal contribution\\
\affilnum{1}Department of Robotics, University of Michigan, Ann Arbor, MI, USA\\
\affilnum{2}Great Lakes Research Center, Michigan Technological University, Houghton, MI, USA}

\corrauth{Advaith Sethuraman, 2505 Hayward St., Ann Arbor, MI 48109}
\email{advaiths@umich.edu}

\begin{abstract}
Open-source benchmark datasets have been a critical component for advancing machine learning for robot perception in terrestrial applications. Benchmark datasets enable the widespread development of state-of-the-art machine learning methods, which require large datasets for training, validation, and thorough comparison to competing approaches. Underwater environments impose several operational challenges that hinder efforts to collect large benchmark datasets for marine robot perception. Furthermore, a low abundance of targets of interest relative to the size of the search space leads to increased time and cost required to collect useful datasets for a specific task. As a result, there is limited availability of labeled benchmark datasets for underwater applications. We present the AI4Shipwrecks dataset, which consists of 28 distinct shipwrecks totaling 286 high-resolution labeled side scan sonar images to advance the state-of-the-art in autonomous sonar image understanding. We leverage the unique abundance of targets in Thunder Bay National Marine Sanctuary in Lake Huron, MI, to collect and compile a sonar imagery benchmark dataset through surveys with an autonomous underwater vehicle (AUV). We consulted with expert marine archaeologists for the labeling of robotically gathered data. We then leverage this dataset to perform benchmark experiments for comparison of state-of-the-art supervised segmentation methods, and we present insights on opportunities and open challenges for the field. The dataset and benchmarking tools will be released as an open-source benchmark dataset to spur innovation in machine learning for Great Lakes and ocean exploration. The dataset and accompanying software are available at \url{https://umfieldrobotics.github.io/ai4shipwrecks/}.
\end{abstract}

\keywords{Marine robotics, side scan sonar, deep learning, segmentation, benchmark datasets}

\maketitle

\section{Introduction}

It is estimated that over $3$ million undiscovered shipwrecks lie on the ocean floor \citep{gonzalez2009unesco}. Locating these submerged archaeological sites enables research into important maritime assets of historical significance. However, searching over large areas and vast depths of the sea requires expensive and time-consuming surveys, which inhibits new discovery of shipwreck sites. Marine robotic platforms, including autonomous underwater vehicles (AUVs), have demonstrated potential to carry out efficient, cost-effective large-area surveys of marine environments returning hundreds of gigabytes worth of data. Still, the interpretation of sonar imagery to identify sites of interest requires manual expert review. This can take many months to complete, often requiring multiple surveys to verify potential new discoveries. 
\begin{figure}[!ht]
\centering
\includegraphics[width=0.68\linewidth]{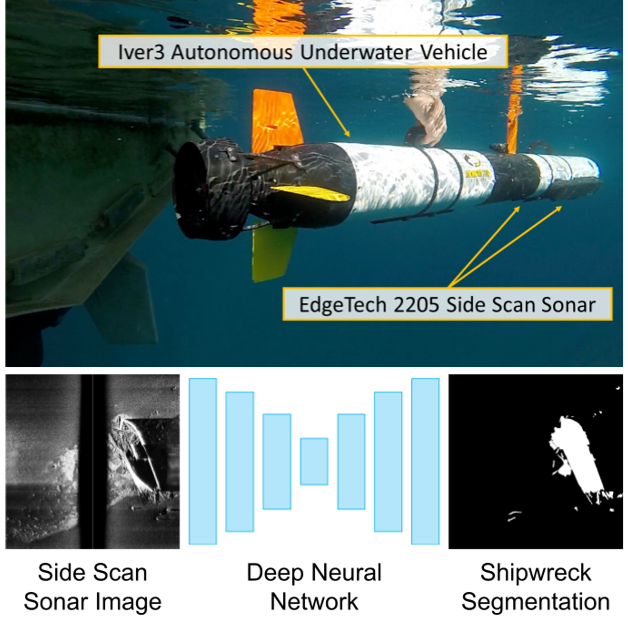}
\vspace{-2mm}
\caption{Our AI4Shipwrecks dataset aims to accelerate the development of shipwreck segmentation algorithms for sonar data collected onboard autonomous systems.\vspace{-9mm}}
\label{fig:intro}
\end{figure}

\begin{table*}[]
\caption{Table comparing existing publicly available sonar datasets for object detection and segmentation (FLS denotes forward looking sonar, and SSS denotes side scan sonar). Note that our dataset is the only one to offer pixel-level segmentation labels for shipwrecks in a real-world environment. \label{data_table}}
\centering
\begin{tabular}{l|ccccc}
\hline
\textbf{Dataset}               & \textbf{Environment} & \textbf{\begin{tabular}[c]{@{}c@{}}Sonar \\ Type\end{tabular}} & \textbf{\begin{tabular}[c]{@{}c@{}}\# Total \\ Images\end{tabular}} & \textbf{\begin{tabular}[c]{@{}c@{}}\# Shipwreck \\ Images\end{tabular}} & \textbf{\begin{tabular}[c]{@{}c@{}}Segmentation\\  Labels\end{tabular}} \\ \hline
Marine Debris FLS (2021)       & Indoor Tank          & FLS                                                            & 1868                                                                & 0                                                                       & \ding{51}                                                                                                                                            \\
SeabedObjects-KLSG (2020)      & Real World           & SSS                                                            & 1190                                                                & 385                                                                     &\ding{55}                                                                                                                                             \\
AURORA (2022)                  & Real World           & SSS  & 2 & 0 & \ding{55} \\
SeabedObjects-KLSG-II (2022)   & Real World           & SSS                                                            & 1296                                                                & 487                                                                     & \ding{55}                                                                                                                                              \\
Marine-PULSE (2023)            & Real World           & SSS                                                            & 719                                                                 & 0                                                                       & \ding{55}                                                                                                                                           \\
Burguera and Bonin-Font (2020) & Real World           & SSS                                                            & 10                                                                  & 0                                                                       & \ding{51}                                                                                                                                           \\ \hline
AI4Shipwrecks (ours) (2024)    & Real World           & SSS                                                            & 286                                                                 & 161                                                                     & \ding{51}                                                                           \\ \hline                                                                 
\end{tabular}
\end{table*}

Automated processing of sonar data collected over large-area surveys has the potential to accelerate the discovery of new sites of interest. On land, machine learning has led to great advances in computer vision and robot perception tasks, including object detection, semantic segmentation, and scene understanding. State-of-the-art machine learning methods rely on labeled datasets for supervised training of neural networks to learn pixel-wise segmentation predictions. However, for underwater domains, there is limited availability of public, labeled datasets for sonar data due to challenges, time, and expense associated with data collection \citep{singh, ochal2020comparison}.

In this paper, we present a new dataset, AI4Shipwrecks, and we present benchmarking results for segmentation of shipwrecks from side scan sonar (SSS) imagery collected on an AUV (Fig.~\ref{fig:intro}). The AI4Shipwrecks dataset contains 286 high-resolution side scan sonar images in .PNG format, with accompanying pixel-wise segmentation labels of shipwreck sites. This dataset was collected over the span of five weeks in Thunder Bay National Marine Sanctuary (TBNMS) in Lake Huron, Michigan, USA. Spanning 4300-square-miles, TBNMS contains almost 100 known shipwreck sites and over 100 undiscovered sites. We leverage the unique abundance of known shipwrecks to curate a rich dataset of 28 distinct shipwrecks for investigating the application of machine learning for this task. In this paper, we present details of dataset collection and preparation for easy indexing of our dataset by machine learning pipelines. To encourage further advances in segmentation for SSS, we report extensive results on our dataset from modern, state-of-the-art segmentation models. Lastly, we include discussion on lessons learned from our field expeditions and experiments to provide insight on future challenges and opportunities for machine learning for processing sonar data. The resulting dataset and benchmarking tools will be made publicly available as a benchmark dataset for segmentation from sonar imagery to enable future research in machine learning for ocean exploration. 

\begin{figure*}
    \centering
\includegraphics[width=0.97\textwidth]{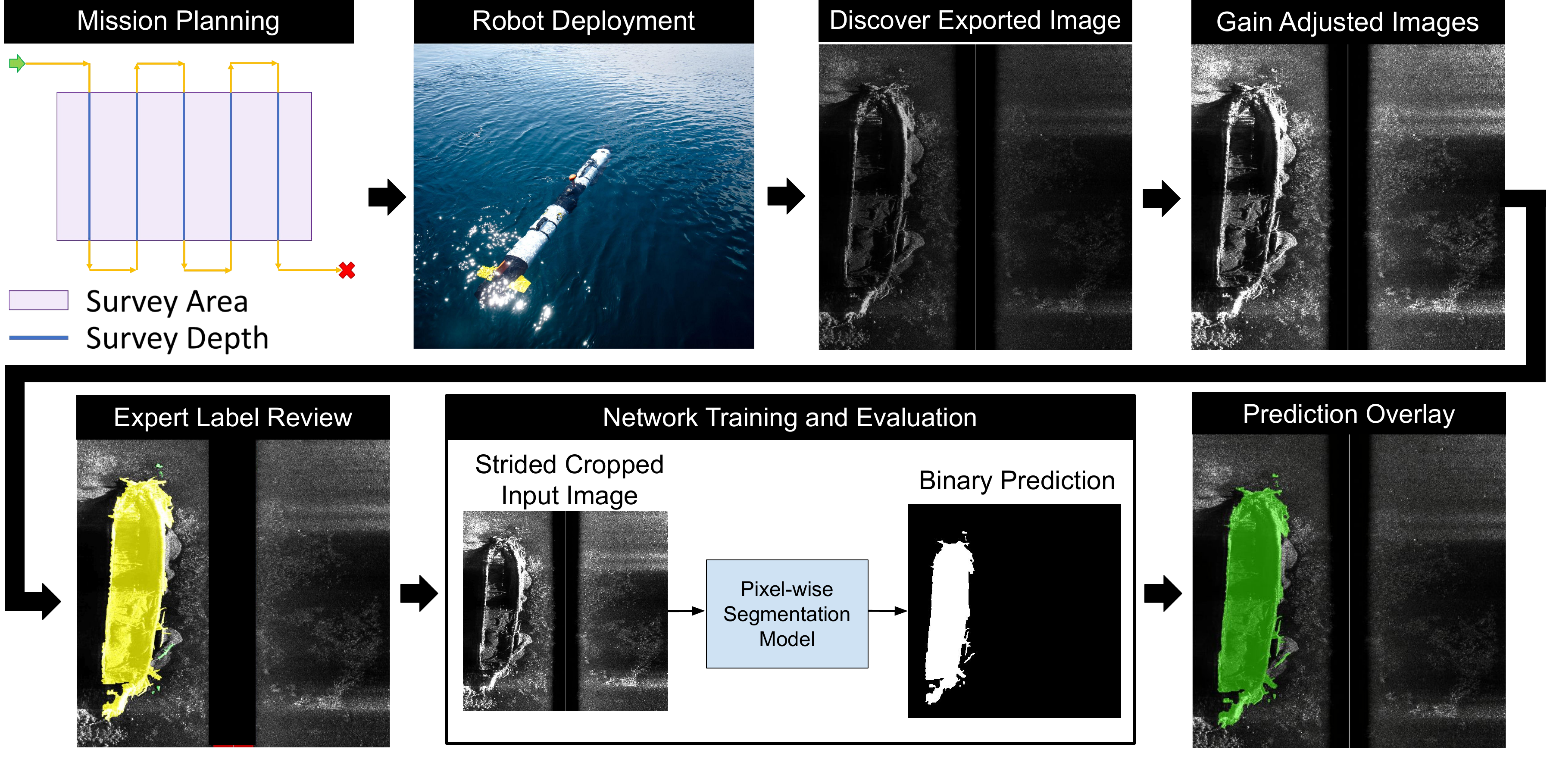}
\vspace{-3mm}
    \caption{Data acquisition, processing, and network inference pipeline using the Iver3 autonomous underwater vehicle. The yellow lines in Mission Planning denote the AUV's trajectory, with blue segments occuring at survey depth.\vspace{-2mm}}
    \label{fig:flowchart}
\end{figure*}

\section{Background}
Sonar is a popular perception sensor underwater due to its long operational range \citep{Lin_2023}.  As a result, sonar image understanding algorithms have the potential to enable autonomous underwater vehicles to operate in unstructured environments.

\subsection{SSS Imagery vs. RGB Imagery}
Converting the raw acoustic data to sonar imagery makes it more easily viewable by humans. However, there are key phenomena found in SSS imagery that are not present in RGB imagery. SSS imagery primarily measures the intensity of sound returned at a given range. Although it may look similar to an optical birds-eye-view of the underwater terrain, the SSS sensor model is dissimilar to RGB camera models. First, the sonar's beam pattern unevenly distributes energy towards the seafloor, leading to varying image intensities along the horizontal axis \citep{jmse8080557}. Next, SSS is a time-of-flight sensor that exhibits acoustic shadows due to occlusions \citep{jmse8080557}.  Finally, the resolution of SSS imagery depends on several parameters such as the altitude of the vehicle, range, and grazing angle, all of which can change within a single image \citep{rs15235599}.

\subsection{Object Segmentation in Sonar Imagery}
 Since many sonar images can be manipulated as single-channel images after post-processing, sonar image understanding algorithms have many similarities to techniques intended for RGB images from computer vision.

Recent work focuses on the application of data-driven deep neural networks for object detection and segmentation in sonar imagery. The majority of work in object detection for sonar imagery involves retraining or fine-tuning existing object detection algorithms on task-specific datasets of sonar imagery \citep{8604879, SSSEG}. The unique waterfall data format of SSS data has also motivated the development of specialized network architectures that enable real-time inference on streamed SSS data \citep{jmse8080557}. 

\subsection{Datasets}
Large, labeled, and publicly available SSS datasets incur high costs of collection and require expert analysis for labeling. As a result, there are fewer options for evaluating the performance of object detection and segmentation algorithms meant for sonars. A summary of existing public sonar datasets for machine learning applications is shown in Table \ref{data_table}.

There has been recent interest in deep learning datasets for forward looking sonars (FLS) \citep{singh, dataset1, Xie_2022, 9721640}. \citeauthor{singh} present an FLS dataset of debris in marine environments, but FLS is shorter range and has distinct sensor geometry compared to SSS. Furthermore, the dataset is object-centric and captured in a controlled lab environment. 

There are two recent SSS datasets that contain submerged objects, namely SeabedObjects-KLSG \mbox{\citep{seabedobject}} and Marine-PULSE \mbox{\citep{Marine_PULSE}}. However, neither of these datasets contain pixel-wise semantic labels. \citeauthor{nnms-te61-20} present SSS and AUV data of terrain in the Greater Haig Fras Marine Conservation Zone, however the dataset does not contain any shipwrecks and also does not have semantic labels. Similarly, the SSS dataset released by \citet{jmse8080557} focuses exclusively on seafloor terrain. While this dataset does include pixel-wise labels, the labels are only for terrain types: ``sand,'' ``rock,'' and ``other.'' There are no shipwrecks included in the surveys and the dataset only contains a total of ten waterfall images.

To the best of our knowledge, our AI4Shipwrecks dataset is the first publicly available dataset for shipwreck segmentation in SSS images in the marine autonomy community.

\begin{figure*}[ht] 
        \centering
        \includegraphics[width=0.97\linewidth]{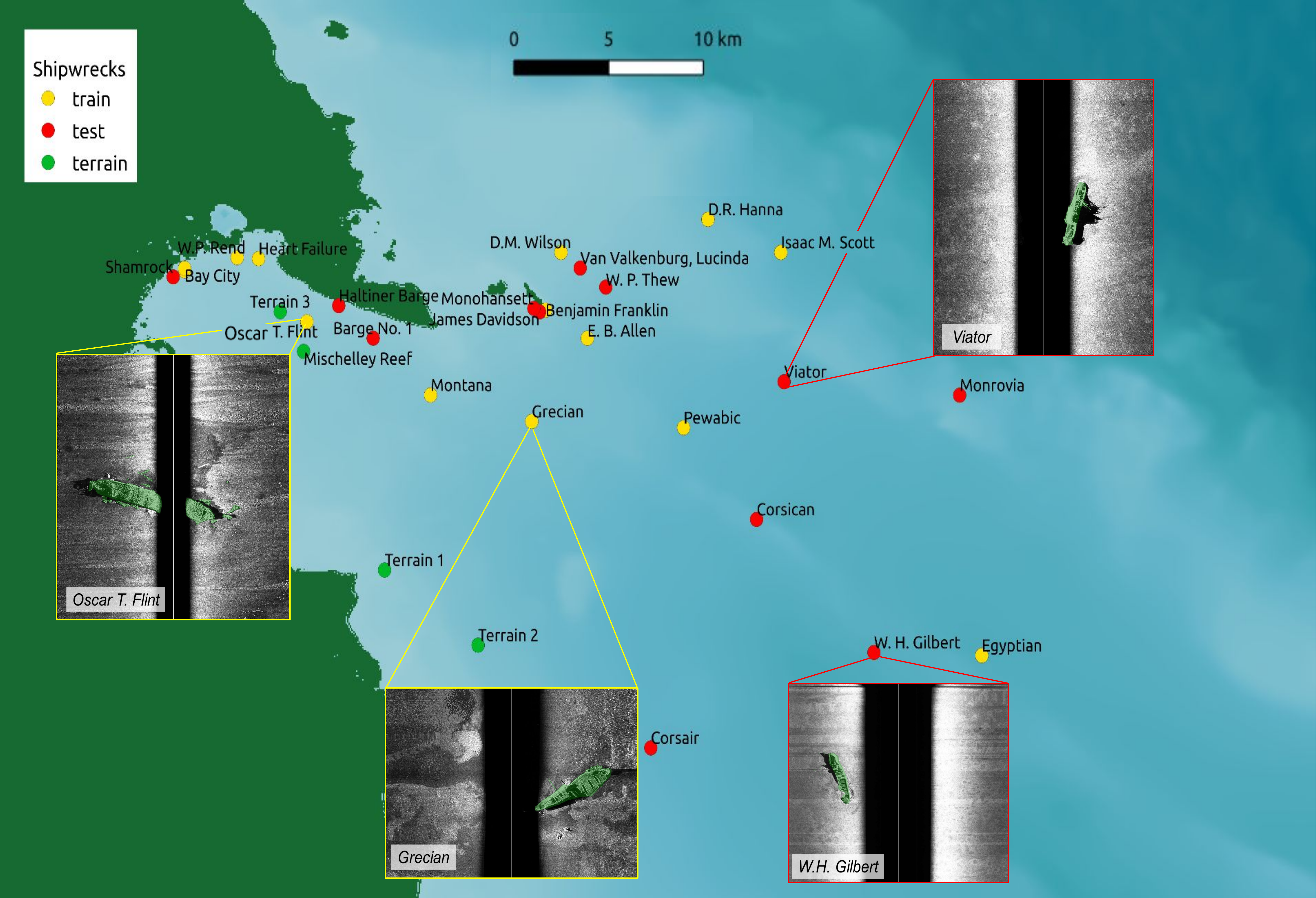}
        \caption{Map of survey sites in TBNMS, Lake Huron, MI. Callouts include example sonar data overlaid with ground truth labels. Color indicates sites that are included in testing (red) and training (yellow) splits, and locations of additional terrain surveys (green). Best viewed in color and zoomed in. \vspace{-5mm}}
        \label{fig:fieldmap}
    \end{figure*}

\subsection{Synthetic Datasets for Sonar}
Simulation can produce low-cost, diverse, and labeled datasets for sonar image understanding \citep{doi:10.1080/01691864.2021.1873845, CERQUEIRA2020101086, sethuraman2023stars, lee2018deep}. Many simulators use ray-tracing techniques to render arbitrary 3D meshes \citep{sethuraman2023stars, CERQUEIRA2020101086}, while others use style-transfer techniques to mimic the noise and sensor models of sonars \citep{lee2018deep, SUNG2019291}. Methods that train solely on simulated data encounter a \textit{sim-to-real gap} between simulated sonar imagery and real imagery, which leads to reduced performance when testing on real data \citep{sethuraman}. Prior work has studied the sim-to-real gap and techniques for reducing it \citep{sethuraman2023stars}. Although simulated data can help mitigate training data scarcity, it is still desirable to \textit{evaluate} the performance of vision algorithms on a real test dataset for deployment in the real world. AI4Shipwrecks addresses this gap by leveraging the natural diversity of shipwreck sites in TBNMS. We present a dataset representative of real shipwreck sites an AUV could encounter during autonomous surveys. We also release pixel-wise segmentation labels of shipwreck sites to enable thorough evaluation of machine learning methods for the task of shipwreck segmentation in real environments.

\section{Technical Approach}
    
Figure~\ref{fig:flowchart} provides an overview of the developed pipeline for acquiring and testing sonar data for machine learning applications. 
First, data is collected through deploying an AUV for a large-area survey. 
The AUV carries out a pre-programmed survey mission to acquire sonar data. 
Once the data are returned, post-processing converts the raw sonar data format (.JSF) to standard image format (.PNG). 
The standard image format is further processed to be input into a deep neural network. 
The network outputs a prediction in the form of a binary per-pixel segmentation mask, which can be visualized as an overlay on the input sonar image.

    \subsection{Site Selection}
    Surveys were conducted in TBNMS in Lake Huron, MI. 
    Figure~\ref{fig:fieldmap} shows shipwreck sites that were imaged in our surveys during five weeks over the course of two years.
    The abundance of known shipwreck targets in TBNMS was a crucial factor in our selection of this field site, as this enabled us to maximize the number of targets observed within a relatively constrained area and short timespan. 
    We timed underwater surveys during late-May through early-June to mitigate the effects of thermocline. We did not conduct surveys in inclement weather.
    These proposed survey regions were selected in coordination with scientists at TBNMS to cover a wide range of ship types, site relief, wreck characteristics, and water depth, prioritizing sites within a reasonable distance from the Port of Alpena. 
    This ensured that we could survey a maximum number of sites while still capturing variation across samples, providing a unique and valuable dataset for training machine learning methods. 
    Furthermore, this large and diverse dataset allows us to validate and thoroughly evaluate the accuracy and generalizability of developed methods.
    
    \subsection{Data Collection Platform}
     \begin{figure}[ht]
        \centering
        \includegraphics[width=0.48\textwidth, trim=3 3 3 3,clip]{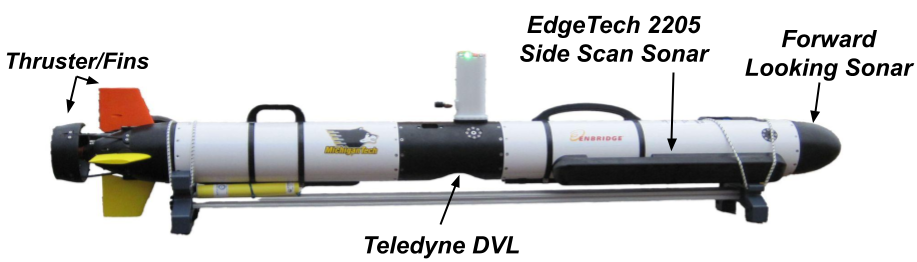}
        \caption{Iver3 data collection platform equipped with advanced localization and high-resolution seafloor mapping capabilities.}
        \label{fig:iver3}
    \end{figure}
  
    The surveys were conducted using the Michigan Technological University (MTU) Iver3 AUV equipped with an EdgeTech $2205$ dual frequency $540/1610$ kHz SSS and 3D bathymetric system (pictured in Fig. \ref{fig:iver3}). Note that the EdgeTech 2205 uses CHIRP and not continuous wave pulses (CW), and as a result there is a CHIRP start frequency ($C_s$) and CHIRP end frequency ($C_e$). Relevant sensor parameters are reported in Table \ref{sonar_params}.
    Each survey mission was conducted by pre-programming a route to survey target sites. 
    The AUV was programmed to capture SSS data while driving at a constant velocity of $2.5$ knots.
    The released imagery is produced from the low frequency sonar, as a lower frequency is able to cover a wider swath of area compared to high frequency sonar. 
    The AUV's Teledyne RDI Explorer Doppler Velocity Log (DVL) provides a long-term accuracy of $\pm0.3 \% \pm0.2 $cm/s \citep{DVL_Manual}. The AUV's DVL has a maximum range of 81 meters \citep{DVL_Manual}. The speed of sound in water is measured by an AML Oceanographic Sound Velocity Sensor, with a range of 1375-1625 m/s, precision of $\pm$ 0.006 m/s, accuracy of $\pm$ 0.025 m/s, and resolution of 0.001 m/s.
    The AUV is also equipped with a forward-looking obstacle avoidance sonar. 

\begin{table}[]
\centering
\caption{Table with side scan sonar parameters. Note sonar range indicates the selected ranges during our surveys, not the min/max ranges for the sensor. The beam width $\theta_h$ is two-way. \label{sonar_params}}
\begin{tabular}{l|c}
\hline
\textbf{Sensor Parameter}      & \textbf{Value }       \\ \hline
Center Frequency ($f$)   & 520 kHz      \\
CHIRP Start Freq. ($C_s$) & 488.5 kHz \\ 
CHIRP End Freq. ($C_e$) & 551.5 kHz \\
Sweep Length ($\tau$)        & 1 ms         \\
Horizontal Beam Width ($\theta_h$) & 0.26 degrees \\
Survey Speed $(V_s)$         & 2.5 knots    \\
Sonar Ranges $(R)$         & 30-150 m  \\
DVL Maximum Range & 81 m \\
DVL Long Term Accuracy & $\pm0.3 \% \pm0.2 $ cm/s \\ 
Sound Velocity Sensor Range & 1375-1625 m/s \\ 
Sound Velocity Sensor Precision &  $\pm$ 0.006 m/s \\ 
Sound Velocity Sensor Accuracy & $\pm$ 0.025 m/s \\ 

\end{tabular}
\end{table}

  \begin{figure*} [ht]
\includegraphics[width=\linewidth]{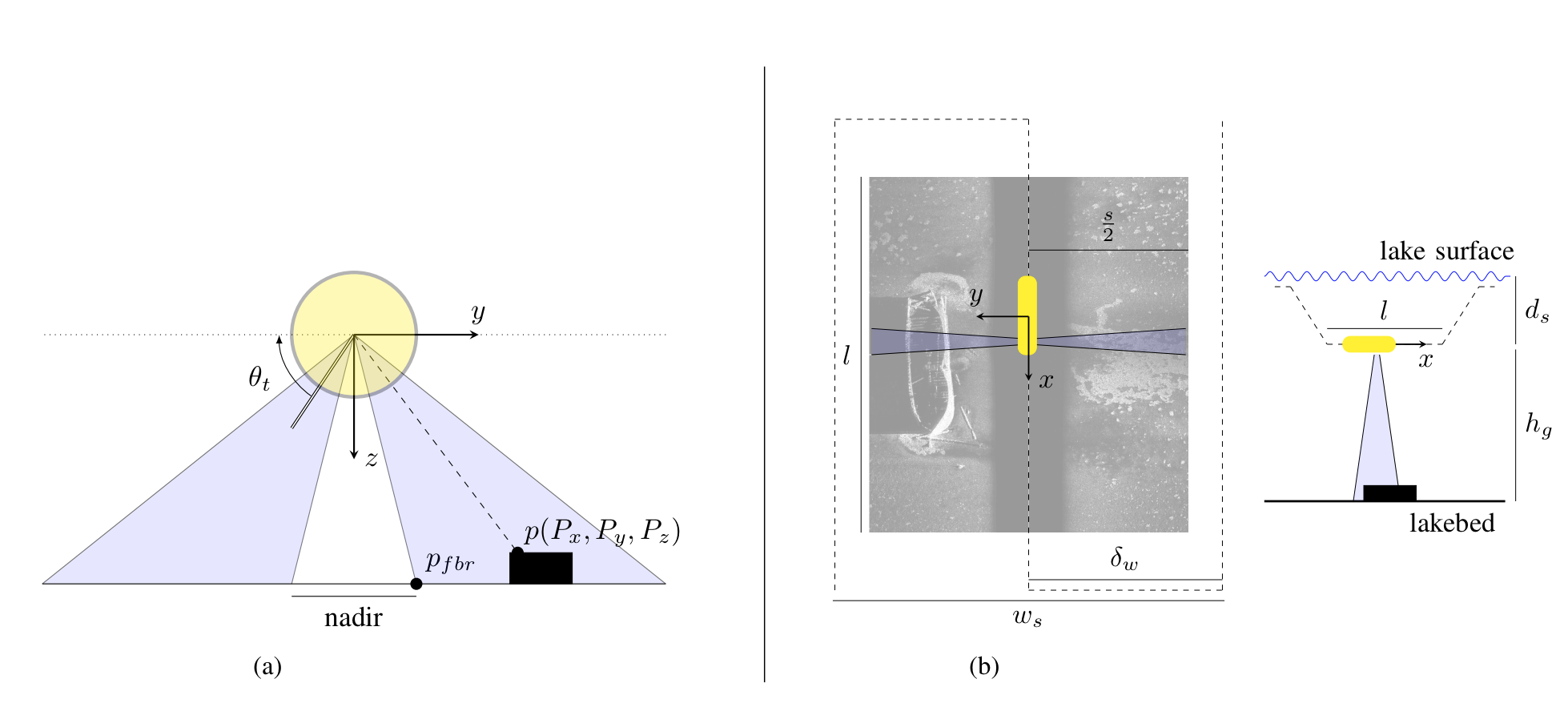}

    \caption{a) SSS sensor model detailing sensor tilt angle ($\theta_t$), first bottom return ($p_{fbr}$), nadir gap, and ensonified point on object ($p(P_x, P_y, P_z)$). The SSS field of view is shown in blue. b) The AUV (in yellow) performs a survey with half-swath width ($\frac{s}{2}$), leg width ($\delta_w$), leg length ($l$), and total survey width ($w_s$). For each survey leg, the AUV dives to the depth from surface ($d_s$) or height over ground ($h_g$) and resurfaces between legs to acquire a GPS update. The SSS is only collecting data once submerged at depth ($d_s$) for length ($l$).}\vspace{-3mm}
    \label{fig:sensor_model}
    
\end{figure*}
    \subsection{Side Scan Sonar Sensor Model}

    The EdgeTech 2205 SSS used in our surveys emits low and high frequency acoustic CHIRPs from two transducers aimed at the seafloor. The signal travels through the water column and reflects off the terrain or other objects in the swath area before being received by the sensor. After a sonar CHIRP has been emitted, the sensor receives and bins the intensity of returns according to time-of-flight. On the horizontal axis (across-track), an SSS image is a histogram of return intensity at equally spaced intervals in time. Each histogram is accumulated in the vertical axis (along-track) as the transducer moves to produce an image with two dimensions. The higher the returned signal intensity, the higher the pixel value in the resulting sonar image. Raw SSS imagery is single channel grayscale but can be viewed with various color palettes for improved visibility. 
    
\begin{figure}
        \includegraphics[width=\linewidth]{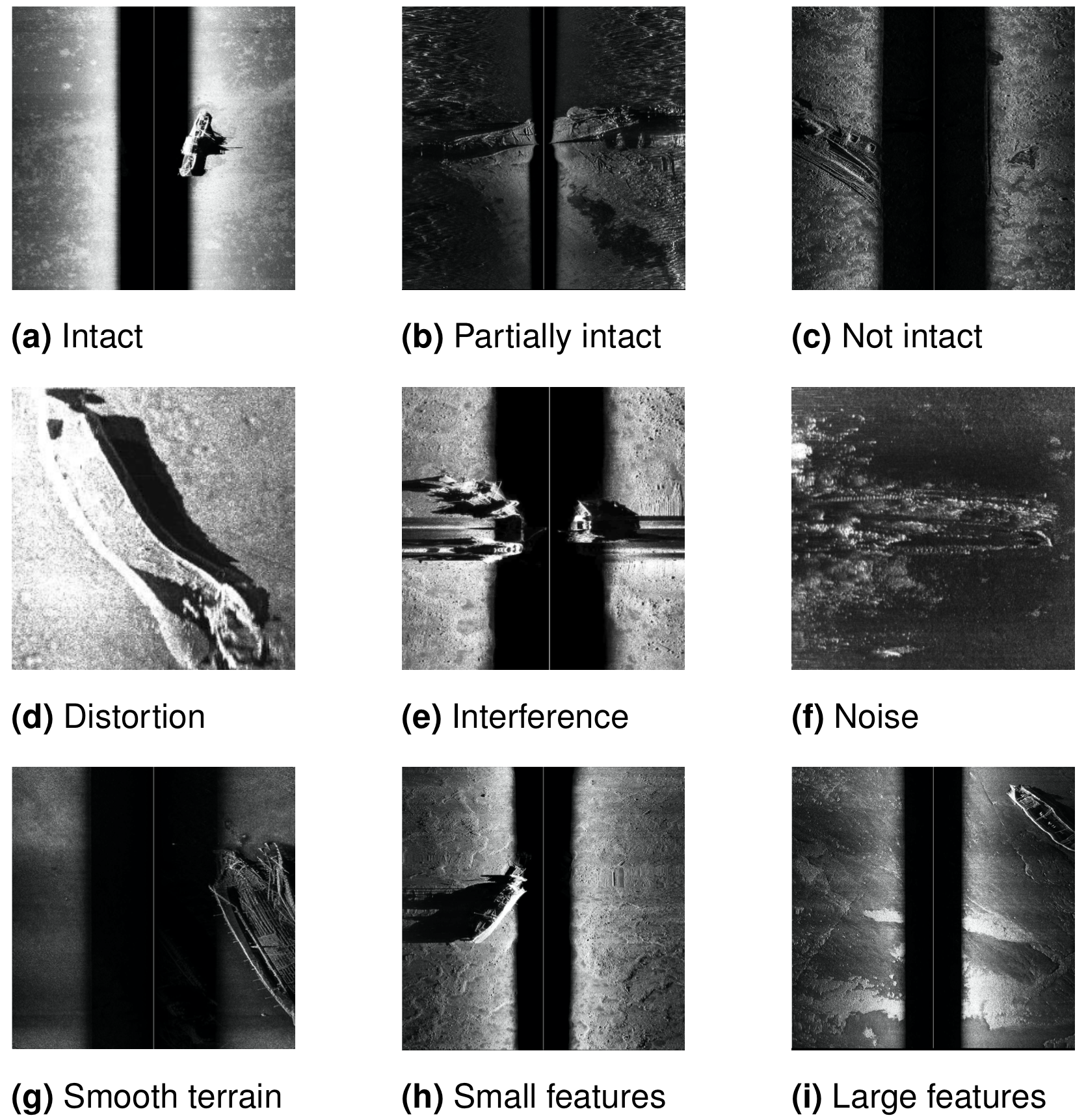}
        \caption{Images of wrecks highlighting levels of the three major categories used to split the dataset. Figures 6a-6c represent the shipwreck conditions, Figs. 6d-6f represent different sonar quality conditions, and Figs. 6g-6i represent the terrain types. \vspace{-5mm}}
        \label{fig:metrics}
    \end{figure}

    The sensor model is depicted in Fig. \ref{fig:sensor_model}a and illustrates an acoustic beam that encounters an object at point $p(P_x, P_y, P_z)$. SSS typically has two identical transducer arrays mounted symmetrically on either side of the AUV at a fixed tilt angle, $\theta_t$. The first return from the sonar is called the first bottom return ($p_{fbr}$) and is used as an estimate of AUV altitude. There is typically a sonar deadzone called a \textit{nadir} due to the transducer beam pattern and geometry of the mounted sensor on the AUV. This causes a black stripe down the center of sonar images, as seen in Fig. \ref{fig:metrics}.
    
    
    Although SSS can produce detailed images of the environment, view-dependent shadowing effects, self-occlusion, material-dependent acoustic noise, and distortion make object detection a difficult task for both humans and automated algorithms. Figures \ref{fig:metrics}d-f illustrate common distortions and noise found in SSS imagery.  

    \subsection{Side Scan Sonar Resolution}
    SSS resolution is dependent on a variety of sensor and environmental parameters \citep{rs15235599}. These include the horizontal beamwidth ($\theta_h$), sonar range ($R$) reported in Table \ref{sonar_params}, and the speed of sound ($c$). Along-track resolution ($R_x$) is defined as the resolution in the direction of vehicle travel (vertical image axis):
    \begin{equation}
        R_x =  R \times sin(\theta_h)
    \end{equation}
    Similarly, across-track resolution ($R_y$) refers to the resolution along the range axis (horizontal image axis). This is calculated based on the bandwidth (BW = $C_s - C_e$) of the sonar CHIRP used. 
    \begin{equation}
    R_y = \frac{c}{2 \times BW}
    \end{equation}
    Using these formulas, we can calculate the resolutions of sonar imagery captured. We report along-track resolutions at the maximum sonar range and across-track resolutions at the measured speed of sound ($c$) in Table \ref{survey_res}. 

    \subsection{AUV Survey Mission Planning}
    The Iver3 AUV has two standard patterns for pre-planned missions: lawnmower (LM) and object identification (OID). The LM pattern is a common down-and-back structure, as depicted by the dotted lines in Fig. \ref{fig:sensor_model}b. In one leg of a LM survey, there are three stages: descent, survey, and ascent. The AUV must ascend, turn around, and then descend before collecting sonar data for each leg in order to geolocalize itself, as global positioning system (GPS) data is not available underwater. The survey occurs at either a constant height from bottom ($h_b$) or depth from surface ($d_s$), and this parameter may change from site to site based on the depth of the wreck. Given that the maximum range of the DVL is 81m, and the maximum depth of the shipwreck sites surveyed was 79.2m, we may conclude that bottom-tracking was not lost during repeated descent/ascent stages. The leg length ($l$) in Fig. \ref{fig:sensor_model}b denotes the distance that the AUV was underwater collecting SSS data, and the half-swath width ($\frac{s}{2}$) describes the distance on the seafloor covered by the sonar beams from one of the transducer arrays. For a planned LM mission, $w_s$ is the total survey width and $\delta_w$ is the width of each individual leg. 
    
            \begin{table*}[]
         \caption{Detailed survey parameter information including sonar range, leg length, survey depth (from surface) or survey height (from bottom). Note that sonar range and leg length are related to the width and height of a captured SSS image in meters respectively. All values were recorded prior to deployment of the AUV and cross-referenced with information from the .JSF files. Note we are describing each individual survey in this table rather than aggregated per site, as multiple surveys at one location may have had different parameters. \label{survey_res}}
         \centering
\begin{tabular}{l|cccccc}
\hline
\textbf{Survey}                                          & \textbf{\begin{tabular}[c]{@{}c@{}}2*Sonar \\ Range (m)\end{tabular}} & \textbf{\begin{tabular}[c]{@{}c@{}}Leg \\ Dist. (m)\end{tabular}} & \textbf{\begin{tabular}[c]{@{}c@{}}Depth\\ (m)\end{tabular}} & \textbf{\begin{tabular}[c]{@{}c@{}}Height\\ (m)\end{tabular}} & \textbf{\begin{tabular}[c]{@{}c@{}}$R_x$ \\ Range (cm)\end{tabular}} & \textbf{\begin{tabular}[c]{@{}c@{}}$R_y$\\ (cm)\end{tabular}} \\ \hline
\textit{20220523-141805--EBAllenLM}                    & 60                                                                    & 110                                                              & 25                                                           & --                                                              & 13.61                                                              & 1.16                                                       \\
\textit{20220523-154917--EBAllenLM15mDFS}              & 60                                                                    & 110                                                              & 15                                                           & --                                                              & 13.61                                                              & 1.15                                                       \\
\textit{20220523-185512--Thew15DFS}                    & 60                                                                    & 150                                                              & 15                                                           & --                                                              & 13.61                                                              & 1.17                                                       \\
\textit{20220523-204910--Thew15DFS54oa}                & 60                                                                    & 150                                                              & 15                                                           & --                                                              & 13.61                                                              & 1.17                                                       \\
\textit{20220524-134748--VanValkenburgLM10mDFS}                 & 60                                                                    & 100                                                              & 10                                                           & --                                                              & 13.61                                                              & 1.17                                                       \\
\textit{20220524-150629--VanValkenburgOID10mDFS}                & 60                                                                    & 150                                                              & 10                                                           & --                                                              & 13.61                                                              & 1.13                                                       \\
\textit{20220524-161604--WilsonOID2mDFS}               & 60                                                                    & 150                                                              & 2                                                            & --                                                              & 13.61                                                              & 1.13                                                       \\
\textit{20220524-174036--WilsonBroadSearch2mDFS}       & 60                                                                    & 1000                                                             & 2                                                            & --                                                              & 13.61                                                              & 1.13                                                       \\
\textit{20220524-201003--MontanaOID8mDFS}              & 60                                                                    & 150                                                              & 8                                                            & --                                                              & 13.61                                                              & 1.13                                                       \\
\textit{20220527-134851--ViatorOID12mDFS}              & 120                                                                   & 400                                                              & 12                                                           & --                                                              & 27.23                                                              & 1.17                                                       \\
\textit{20220527-180547--MonroviaOID20mDFS}            & 250                                                                   & 400                                                              & 20                                                           & --                                                              & 56.72                                                              & 1.17                                                       \\
\textit{20220527-211838--HaltinerLM2mDFS}              & 250                                                                   & 150                                                              & 2                                                            & --                                                              & 56.72                                                              & 1.15                                                       \\
\textit{20220531-132042--Flint}                        & 120                                                                   & 400                                                              & 2.5                                                          & --                                                              & 27.23                                                              & 1.15                                                       \\
\textit{20220531-145419--Flint\_terrain}               & 240                                                                   & 400                                                              & 2.5                                                          & --                                                              & 54.45                                                              & 1.16                                                       \\
\textit{20220531-162857--Flint\_120m}                  & 240                                                                   & 400                                                              & 2.5                                                          & --                                                              & 54.45                                                              & 1.16                                                       \\
\textit{20220531-180338--Heart\_Failure}               & 260                                                                   & 200                                                              & 1                                                            & --                                                              & 58.99                                                              & 1.17                                                       \\
\textit{20220601-143442--BargeNo1}               & 250                                                                   & 600                                                              & 2                                                            & --                                                              & 56.72                                                              & 1.16                                                       \\
\textit{20220601-171552--Rend}                         & 120                                                                   & 200                                                              & 1                                                            & --                                                              & 27.23                                                              & 1.16                                                       \\
\textit{20220601-185052--Heart\_Failure2}              & 120                                                                   & 200                                                              & 2                                                            & --                                                              & 27.23                                                              & 1.13                                                       \\
\textit{20220602-131155--Grecian}                      & 260                                                                   & 250                                                              & 5                                                            & --                                                              & 58.99                                                              & 1.15                                                       \\
\textit{20220602-141250--Pewabic}                      & 240                                                                   & 400                                                              & 22                                                           & --                                                              & 54.45                                                              & 1.15                                                       \\
\textit{20230605-141930--Egyptian1\_100m\_DFS45m\_LM}  & 200                                                                   & 400                                                              & 45                                                           & --                                                              & 45.38                                                              & 1.14                                                       \\
\textit{20230605-173643--Egyptian2\_130m\_DFS32m\_OID} & 260                                                                   & 300                                                              & 32                                                           & --                                                              & 58.99                                                              & 1.14                                                       \\
\textit{20230605-193324--Gilbert1\_130m\_DFS15m\_LM}   & 260                                                                   & 200                                                              & 15                                                           & --                                                              & 58.99                                                              & 1.13                                                       \\
\textit{20230605-202803--Gilbert2\_160m\_DFS25m\_LM}   & 320                                                                   & 400                                                              & 25                                                           & --                                                              & 72.61                                                              & 1.13                                                       \\
\textit{20230606-181949--NellieGardner1\_HFB\_LM}      & 120                                                                   & 400                                                              & --                                                           & 3                                                             & 27.23                                                              & 1.13                                                       \\
\textit{20230606-195902--NellieGardner2\_HFB\_LM}      & 70                                                                    & 400                                                              & --                                                           & 3                                                             & 15.88                                                              & 1.13                                                       \\
\textit{20230607-122726--Monohansett1\_HFB\_LM}        & 100                                                                   & 800                                                              & --                                                           & 3                                                             & 22.69                                                              & 1.13                                                       \\
\textit{20230607-144904--MischelleyReef1\_3mDFS\_LM}   & 80                                                                    & 500                                                              & 3                                                            & --                                                              & 18.15                                                              & 1.13                                                       \\
\textit{20230607-164552--MischelleyReef2\_3mDFS\_LM}   & 80                                                                    & 400                                                              & 3                                                            &  --                                                             & 18.15                                                              & 1.13                                                       \\
\textit{20230607-175515--NearShore1\_2mHFB\_LM}        & 60                                                                    & 500                                                              & 3                                                            &  --                                                             & 13.61                                                              & 1.14                                                       \\
\textit{20230608-125411--Hanna1\_100m\_DFS20m\_LM}     & 300                                                                   & 600                                                              & 10                                                           &  --                                                             & 68.07                                                              & 1.14                                                       \\
\textit{20230608-154854--Terrain1\_3mDFS\_LM}          & 160                                                                   & 2000                                                             & 2.5                                                          &  --                                                             & 36.30                                                              & 1.14                                                       \\
\textit{20230609-130155--Corsican1\_100m\_DFS5m\_LM}   & 300                                                                   & 400                                                              & 5                                                            &  --                                                             & 68.07                                                              & 1.14                                                       \\
\textit{20230609-155043--Corsican2\_100m\_DFS5m\_LM}   & 300                                                                   & 700                                                              & 5                                                            &  --                                                             & 68.07                                                              & 1.14                                                       \\
\textit{20230609-173705--Corsair1\_700m\_DFS12m\_LM}   & 300                                                                   & 500                                                              & 12                                                           &  --                                                             & 68.07                                                              & 1.14                                                       \\
\textit{20230609-194421--Exploratory1\_15mDFS\_LM}     & 160                                                                   & 2000                                                             & 15                                                           &  --                                                             & 36.30                                                              & 1.14                                                       \\
\textit{20230614-121856--Rend}                         & 80                                                                    & 400                                                              & 1                                                            &  --                                                             & 18.15                                                              & 1.16                                                       \\
\textit{20230614-155731--Monohansett1\_HFB\_LM}        & 70                                                                    & 300                                                              & --                                                           & 3                                                             & 15.88                                                              & 1.14                                                       \\
\textit{20230614-173130--Davidson}                     & 70                                                                    & 420                                                              & 1                                                            &  --                                                             & 15.88                                                              & 1.14                                                       \\
\textit{20230615-124230--Scott1}                       & 300                                                                   & 600                                                              & 15                                                           &  --                                                             & 68.07                                                              & 1.16                                                       \\
\textit{20230615-140855--Scott2}                       & 300                                                                   & 400                                                              & 10                                                           &  --                                                             & 68.07                                                              & 1.16                                                       \\
\textit{20230615-152709--Wilson-2mDFS}                 & 120                                                                   & 600                                                              & 3                                                            &  --                                                             & 27.23                                                              & 1.16                                                       \\
\textit{20230615-165126--VanValkenburg3mDFS}                    & 240                                                                   & 600                                                              & 3                                                            &  --                                                             & 54.45                                                              & 1.17                                                       \\
\textit{20230616-125457--Shamrock1\_1m\_DFS}           & 180                                                                   & 300                                                              & 1                                                            &  --                                                             & 40.84                                                              & 1.16                                                       \\
\textit{20230616-133407--Shamrock2\_1m\_DFS\_2}        & 140                                                                   & 300                                                              & 1                                                            &  --                                                             & 31.76                                                              & 1.16                                                       \\
\textit{20230616-143922--BargeNo1\_23}           & 220                                                                   & 600                                                              & 2                                                            &  --                                                             & 49.92                                                              & 1.16                                                       \\
\textit{20230616-155616--BargeNo1\_Terrain}             & 200                                                                   & 600                                                              & 3                                                            &  --                                                             & 45.38                                                              & 1.16                                                       \\
\textit{20230616-173938--Haltiner\_Terrain}            & 70                                                                    & 600                                                              & 1.5                                                          &  --                                                             & 15.88                                                              & 1.16                                                       \\
\textit{20230616-193812--Haltiner\_Bilge}              & 100                                                                   & 75                                                               & 1.5                                                          &  --                                                             & 22.69                                                              & 1.16                                                      
\end{tabular}
\vspace{15mm}
\end{table*}

    \subsection{Post-Processing} \label{ss:post-processing}
        The Iver3 AUV stores each leg of a sonar survey separately as a .JSF file, which is a proprietary format \citep{jsf}. These files are readable by the free Edgetech Discover software. 
        We used the Discover software for limited sonar image processing and exporting to .PNG. 
        The Discover software normalizes the sonar image intensity and applies Time Varying Gain (TVG). 
        As sound moves through water, absorption loss and spreading cause decreased signal strength. As a result, the returned echoes from objects will also have reduced intensity. TVG addresses this shortcoming and ensures the intensity of the return on the sonar image is not range-dependent \citep{maclennan}. We do not perform slant-range correction within the Discover software \citep{slant-range}. No offline post-processing of the .JSF data was performed using GPS information.

        Pixel-wise labeling of the post-processed sonar images was conducted by a team of three researchers with shared labeling guidelines.
        The labels were then reviewed in-depth by a marine archaeologist from the State of Michigan who is an expert on the shipwrecks at TBNMS. 
        There are two labels: ``shipwreck" and ``other." 
        ``Shipwreck" consists of the primary wrecks as well as any debris.
        If part of the ship is obscured in an acoustic shadow, the label was extrapolated into the shadowed region to follow the expected shape of the wreck based off of expert knowledge.
        The labels are exported to a standard binary mask format where \texttt{0} represents the ``other" class and \texttt{1} represents the ``shipwreck" class.

        As survey lengths ($l$, in meters) can vary for each mission, the resulting sonar images are of different heights ($h$, in pixels). The dataset provides full-sized images of dimension $h \times w$, where $w$ is image width in pixels. We square-cropped these images into overlapping tiles to be input to deep neural networks. 
        
\begin{figure}[ht]
            \centering
            \includegraphics[width=0.75\linewidth]{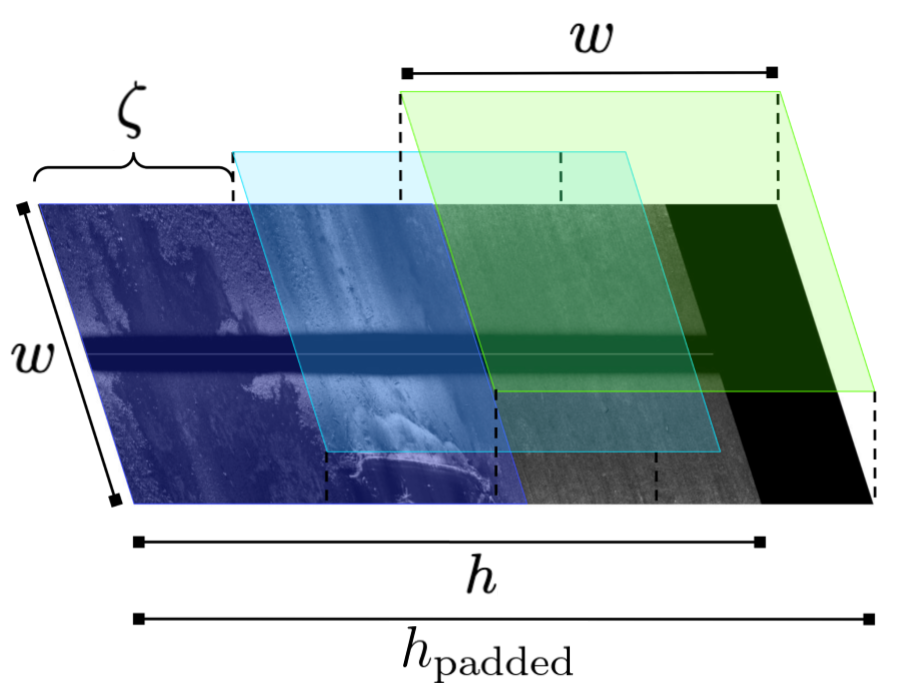}
            \caption{Visualization of the strided cropping of the original side scan sonar image to process before inputting into neural networks. Each colored square represents a $w \times w$ crop of the image, with a stride length of $\zeta=100$ px.}
            \label{fig:stided_crop}
        \end{figure}

        Images are generated by padding the full-size images with black pixels from $h \times w$ to $h_{\text{padded}} \times w$, according to
        \begin{equation}
            h_{\text{padded}} = h + \zeta - [(h - w) \bmod \zeta]
        \end{equation}
        \noindent where $\zeta = 100$ is the stride length in pixels.
        Then, images of dimensions $w \times w$ are cropped every $\zeta$ pixels down the length of the padded image ($h_{\text{padded}}$).
        The image width $w$ output by the Discover software is $1728$ pixels, so the resulting square cropped images are of dimension $1728 \times 1728$.
        A visualization of this process is shown in Fig.~\ref{fig:stided_crop}. The only other data processing step applied to the images is a normalization around 0 based on the mean and standard deviation of the pixel values of the entire dataset.

\section{Dataset Organization}
The dataset is organized by the SSS image and their corresponding labels.
We provide the labels as segmentation masks of objects belonging to a shipwreck in the SSS image, including the shipwreck itself and any debris associated with the shipwreck.
This includes parts of the shipwreck that may have come off from the shipwreck and parts of the ship that have been dislodged. The images and labels are separated into test and train sets using a 50/50 split.
We select a 50/50 split to ensure that the test set is large enough to capture the variance of the shipwrecks sites, ensuring that the robustness of deep learning models can be evaluated through our dataset.
In addition, the images are grouped on a per-site basis.
This is done to ensure that SSS images that see the same terrain, shipwreck, or debris are not shared between the test and train sets of our dataset. \\
\vspace{-5mm}

\begin{figure}[h]
    \centering
    \includegraphics[width=0.7\linewidth]{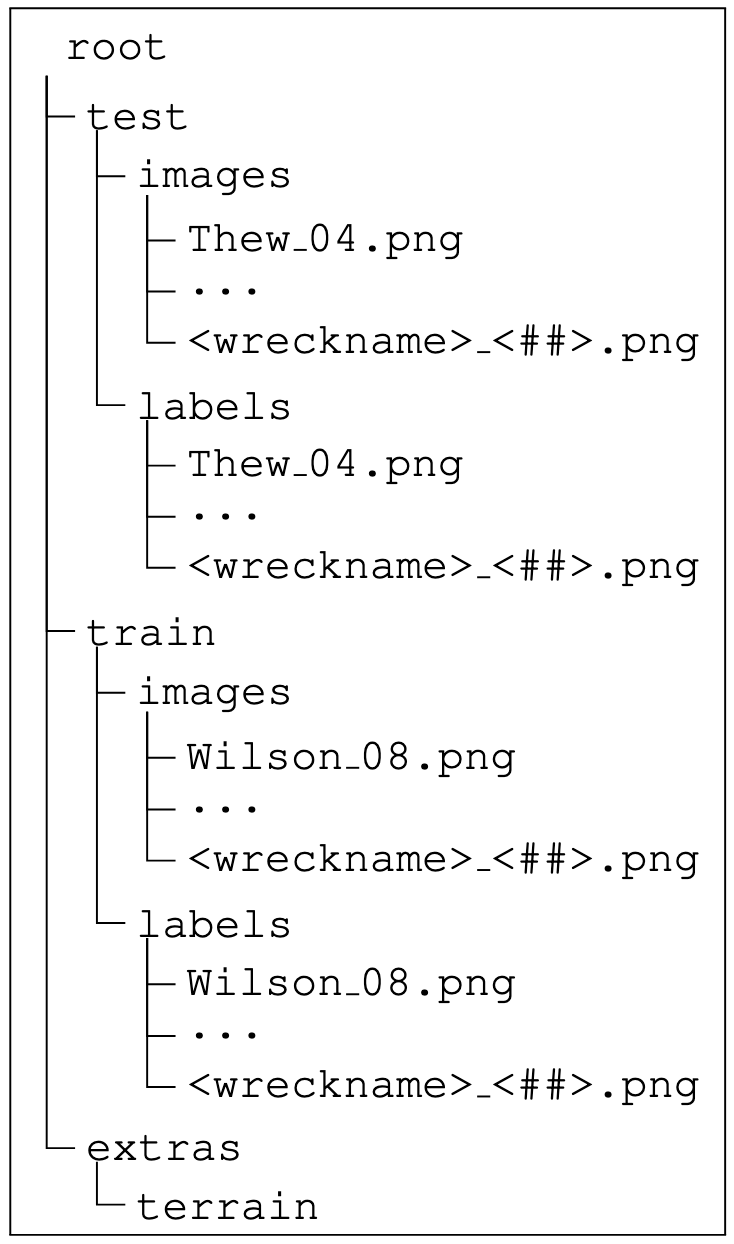}
    \caption{Directory structure of released dataset. Sonar images are sorted into test and train, and named according to their site.\label{fig:tree}}
\end{figure}

\begin{table*}
 \scalebox{0.8}
        \centering
          \caption{Details of site characteristics, image quality, and number of images per site. Depth refers to the distance from the lake surface to the seafloor at the shipwreck site. Condition is in increasing order with 8 corresponding to most intact. SSS quality is in increasing order with 3 corresponding to highest quality. Terrain corresponds to categorical terrain types as defined in Section 4.2.3. $^{*,\dag,\ddag}$ denote shipwrecks that were imaged in a single survey and thus considered as a single ``site." Number of images refers to the number of survey legs collected for each site, where each survey leg is processed into one sonar image. Site properties including depth, length, and beam are obtained from~\cite{tbnms}. We do not present length and beam information for \textit{Heart Failure} because the wreck is very dispersed. Although the main feature of Artificial Reef is not a wreck, there is one wreck imaged in a single scan at that site, whose length and beam information is unknown. Length and beam information for the \textit{Alpena Steamer} is also not known.} 
        \begin{tabu}{l|[2pt]c c c c c c c c}
            \hline
            Name & Depth (m) & Length (m) & Beam (m) & Material & Condition & SSS Quality  & Terrain & \# Imgs. \\
             \hline
              \textbf{Training}   &  &  & &  & &  &  &\\  
            \hline
            \textit{Alpena Steamer$^{\dag}$} &  $3.4$  & -- & -- & Wood & $4$ & $2$ & C & 10 \\
            \textit{Bay City$^{\dag}$} &  $3.4$  & $44.5$ & $8.8$ & Wood & $4$ & $2$ & C & 10 \\
            \textit{D.M. Wilson} & $12.2$ & $54.6$ & $9.8$ & Wood & $6$ & $3$ & B & 22 \\
            \textit{D.R. Hanna} & $41.1$ & $162.2$ & $17.1$ & Steel & $8$ & $2$ & A & 5 \\
            \textit{E.B. Allen}   & $30.5$ & $40.8$ & $7.9$ & Wood & $6$ & $3$ & D & 24 \\
            \textit{Egyptian} & $76.8$ & $70.7$ & $11.0$ & Wood & $5$ & $2$ & D  & 15 \\
            \textit{Grecian} & $30.5$ & $90.2$ & $12.2$   & Steel & $7$ & $2$ & B & 5 \\
            \textit{Harvey Bissell$^{\dag}$} &  $4.6$  & $49.4$ & $10.1$ & Wood & $4$ & $2$ & C & 10 \\
            \textit{Heart Failure} &  $5.5$  & -- & -- & Wood & $2$ & $1$ & B & 12 \\
            \textit{Isaac M. Scott}   & $53.3$  & $159.7$ & $16.5$ & Steel & $8$ & $2$ & C & 6 \\
            \textit{Montana} & $19.2$ & $71.9$ & $11.0$ &  Wood & $6$ & $3$ & A & 8 \\
            \textit{Oscar T. Flint} & $9.1$ & $66.4$ & $11.3$ & Wood &  $5$ & $2$ & A & 22 \\
            \textit{Pewabic} & $50.3$ & $61.0$ & $9.4$ & Wood & $6$ & $2$ & A & 5 \\
            \textit{W.P. Rend} & $5.2$ & $87.5$ & $12.2$ & Wood & $5$ & $3$ & A & 11 \\
            \hline
            \textbf{Testing}   &  &  &  & &  & &  &  \\
            \hline
            Artificial Reef &  $3.0$  & -- & -- & -- & $1$ & $1$ & B & 6 \\
            \textit{Barge No. 1}   & $12.8$ & $94.2$ & $13.4$ & Wood & $5$ & $2$ & A & 15 \\
            \textit{B. Franklin$^{\ddag}$}   & $4.6$ & $41.1$ & $5.8$ & Wood & $2$ & $1$ & B & 10 \\
            \textit{Corsair} &  $55.5$  & $40.8$ & $7.3$ & Wood & $5$ & $2$  & C & 4 \\
            \textit{Corsican} & $48.8$ & $34.1$ &  $7.6$  & Wood & $5$ & $1$ & C & 6 \\
            \textit{James Davidson$^{\ddag}$} & $10.7$   & $30.5$ & $9.1$ & Wood & $4$ & $1$ & A & 10 \\
            \textit{John F. Warner$^{*}$}   & $2.7$ & $38.4$ & $7.9$ & Wood & $3$ & $1$ & B & 6 \\
            \textit{Haltiner Barge}   & $5.2$ & $24.4$ & $10.1$ & Wood & $3$ & $2$ & A & 13 \\
            \textit{L. van Valkenburg}   & $18.3$ & $39.0$ & $7.9$ & Wood & $5$ & $3$ & B & 19 \\
            \textit{Monohansett$^{\ddag}$} & $5.5$ & $48.8$ & $9.1$ & Wood & $5$ & $1$ & B & 10 \\
            \textit{Monrovia} & $42.7$ & $136.6$ & $17.1$ & Steel & $8$ & $3$  & A & 8 \\
            \textit{Shamrock$^{*}$} & $3.4$ & $44.5$ & $9.1$ & Wood & $2$ & $1$ & B & 6 \\
            \textit{W.H. Gilbert} &  $77.7$  & $100.0$ & $12.8$ & Steel & $8$ & $3$ & C & 6 \\
            \textit{W.P. Thew} & $25.6$ & $40.2$ & $7.3$ & Wood & $5$ & $3$ & A & 17 \\
            \textit{Viator} & $57.3$ & $70.7$ & $10.1$ & Steel & $8$ & $2$ & C & 11 \\
            \hline 
        \end{tabu}
        \label{tab:ship_info}
    \end{table*}

\subsection{Released Dataset File Structure}

In Fig. \ref{fig:tree}, the released data directory structure is shown. The data is split into test and train, according to our iterative approach as described in this paper. Within the test and train directories, the sonar images and their corresponding binary labels are stored. Each sonar image (.PNG format) corresponds to one leg of a survey mission. The site of the survey is indicated in the image file names. We also include a directory with four sites of just surveyed terrain in the \texttt{extras} directory. The raw .JSF sonar files -- which contain AUV GPS and positioning data -- will be released on the NOAA National Centers for Environmental Information database and linked to the project webpage.

    \subsection{Train-Test Split Selection}\label{section:dataset_organization.test-train_split_selection}
   
    Table~\ref{tab:ship_info} provides details of each site, organized by train and test split. Our goal with the train-test split is to ensure that each set provides enough diversity of data for both the training and evaluation of deep learning models.
    We inform our split selection through both expert-informed and data-driven metrics, with the target of capturing the underlying distribution of the data as uniformly as possible.
    The atomic unit we use to split up the dataset into train and test sets is each \textit{survey} site location. This was to ensure that no sonar images overlapping the same region are present in both the test and train sets. Please note in Table \ref{tab:ship_info} some sites contain multiple shipwrecks, as indicated by the superscripts.
    To inform the train-test split selection, we group the sites based on three major categories: shipwreck condition, sonar image quality, and terrain type. For each site, a vector score of length three is assigned based on these quantitative and qualitative metrics. 
    A 50/50 train-test split is then performed using an iterative stratification on our multi-labeled data \citep{sechidis}. 

       

        \subsubsection{Wreck Condition:}
            We leverage domain expert knowledge from scientists at TBNMS in order to assign wreck condition labels. Experts who have conducted dive surveys to inspect these ships provided a scale of 1-8 for wreck condition. There are five sites labeled 8 for complete and intact, one site labeled 7 for complete but collapsed, four sites labeled 6 for semi-complete, nine sites label 5 for partially intact, four sites labeled 4 for fragmented, two sites labeled 3 for fragmented and disarticulated, three sites labeled 2 for widely dispersed debris fields, and one site labeled 1 for non-shipwreck cultural materials.

    \begin{table*}[bp] \small
    \vspace{0.5em}
        \renewcommand\thetable{6}
        \caption{Per-site IoU$_{ship}$ compared across each baseline for each site. Higher is better ($\uparrow$).}
        \centering

        \begin{tabu}{c|[2pt]cccccccccccccc}
        \hline
          & \hspace{-1.5em}\parbox[t]{2mm}{\rotatebox[origin=c]{45}{Artificial Reef}}  & \hspace{-1.5em}\parbox[t]{2mm}{\rotatebox[origin=c]{45}{\textit{Barge No. 1}}} & \hspace{-1.5em}\parbox[t]{2mm}{\rotatebox[origin=c]{45}{\textit{Corsair}}} & \hspace{-1.5em}\parbox[t]{2mm}{\rotatebox[origin=c]{45}{\textit{Corsican}}} & \hspace{-1.5em}\parbox[t]{2mm}{\rotatebox[origin=c]{45}{\textit{Davidson}}} &\hspace{-1.5em}\parbox[t]{2mm}{\rotatebox[origin=c]{45}{\textit{W.H. Gilbert}}} & \hspace{-1.5em}\parbox[t]{2mm}{\rotatebox[origin=c]{45}{\textit{Haltiner Barge}}} & \hspace{-1.5em}\parbox[t]{2mm}{\rotatebox[origin=c]{45}{\textit{L. Van Valkenburg}}} & \hspace{-1.5em}\parbox[t]{2mm}{\rotatebox[origin=c]{45}{\textit{Monohansett}}} & \hspace{-1.5em}\parbox[t]{2mm}{\rotatebox[origin=c]{45}{\textit{Monrovia}}} & \hspace{-1.5em}\parbox[t]{2mm}{\rotatebox[origin=c]{45}{\textit{Shamrock}}} & \hspace{-1.5em}\parbox[t]{2mm}{\rotatebox[origin=c]{45}{\textit{W.P. Thew}}} & \hspace{-1.5em}\parbox[t]{2mm}{\rotatebox[origin=c]{45}{\textit{Viator}}}  \\
          \hline
          Yang et. al &0.002&0.471&0.231&0.056&0.000&0.276&0.031&0.367&0.020&0.478&0.102&0.161&0.563\\
          ViT-Adapter&0.003&0.412&0.542&0.088&0.000&0.006&0.368&0.127&\textbf{0.496}&0.469&0.354&0.176&0.631 \\
        DeepLabv3&0.003&0.670&0.531&0.042&0.031&0.658&\textbf{0.459}&0.645&0.034&0.411&0.151&0.419&0.667 \\

            HRNet&0.011&0.703&0.293&\textbf{0.174}&\textbf{0.067}&0.641&0.448&0.641&0.207&0.566&0.003&0.437&0.646  \\
           UNet&0.006&0.736&0.564&0.116&0.074&0.726&0.242&0.641&0.042&\textbf{0.572}&\textbf{0.428}&0.545&0.646  \\
            SOD&\textbf{0.017}&\textbf{0.775}&\textbf{0.583}&0.077&0.039&\textbf{0.749}&0.442&\textbf{0.713}&0.467&\textbf{0.572}&0.001&\textbf{0.580}&\textbf{0.776} \\
          \hline
        \end{tabu}
        \label{tab:metrics_per_site}
    \end{table*}
    
        \subsubsection{Sonar Image Quality Assessment:}
            The sonar image quality assessment is done based on well-developed metrics for optical image quality assessment (IQA). 
            We utilize the no-reference, completely blind IQA metric Natural Image Quality Evaluator (NIQE), which constructs a quality score from a natural scene statistic model trained on undistorted images \citep{mittal}. Although there are some sonar quality metrics \citep{liu-siq, chen-siq}, they generally rely on reference images and measure signal degradation during transmission, which is not suitable for our dataset.
            There are many challenges in evaluating the quality of a sonar image, the first of which being that many IQA metrics are targeted at measuring quality deterioration from compression. 
            However, our sonar images are not compressed -- rather, any distortion in the images is a result of when the Iver3 is surveying shallower sites, making it more vulnerable to drift from waves (Fig. \ref{fig:metrics}d).
            If the shipwreck partially falls into the nadir, this can cause significant interference and acoustic shadows that disrupt the geometry of the wreck (Fig. \ref{fig:metrics}e).
            Additionally, if the shipwreck falls into the outer edges of the sonar image, the increased noise of more distant acoustic returns causes significant noise in the final sonar image (Fig. \ref{fig:metrics}f).
            Informed by the NIQE score, we assign classes of 1-3 to the images where 1 indicates poorer sonar image quality and 3 indicates higher image quality.

                 
         

        \subsubsection{Terrain Type Assessment:}
            We select the terrain type as a category for determining the train-test split, as the varying conditions of terrain can play a crucial role in the detection and segmentation of shipwrecks.
            This is most prominent in terrains that have a large amount of texture, making it difficult even for humans to discern the shipwreck from the terrain, leading to either false negatives or false positives.
            We hence aim to have the terrains split evenly based on their characteristics observed from the SSS imagery.
            In order to determine the most even split across the data based on terrain, we utilize clustering methods.

            Specifically, the clustering is done on the N-dimensional latent space of a Variational Auto Encoder (VAE) pre-trained on a large color image (RGB) dataset.
            The input images are pre-processed by applying white balancing.
            This is done in order to prevent the clustering based on image intensity; otherwise, the intensity differences across the dataset are too prominent.
            We use $k$-means clustering, yielding a partitioning on an image-to-image basis.
            Each image is grouped based on the site it belongs to, which is then used to inform the train-test split. 
            We assign sites a class of A for terrains with small-scale texture, a class of B for terrains with large-scale texture, a class of C for smooth terrain, and a class of D for patchy terrain texture.

    \subsection{Dataset Statistics}
        The dataset consists of $286$ images exported from the Discover software as high-resolution .PNG images. Out of these $286$ images, $161$ contain shipwrecks. There are $1.49$e${7}$ shipwreck pixels and $4.39$e${11}$ background pixels.

\renewcommand\thetable{5}
        \begin{table}[hb]\normalsize
        \centering
         \caption{Aggregate baseline performance averaged across sites: metrics are computed per site and then the average is taken across all test sites for each metric. Metrics include IoU$_{ship}$, F1 Score, TPR, and TNR. TPR and TNR are calculated assuming the \textit{shipwreck} class is positive. $\uparrow$ indicates higher is better.}
         \scalebox{0.86}{
        \begin{tabu}{c| c  c c c }
            \hline
            Baseline     & IoU$_{ship}$ $\uparrow$   & F1 Score $\uparrow$  & TPR $\uparrow$ & TNR $\uparrow$ \\
            \hline
            Yang et. al&	0.212&		0.310&	0.296&	0.995	 \\
            ViT-Adapter&	0.283&		0.395&	0.488&	0.995	             \\
            DeepLabv3 &	0.363&		0.473&	0.485	& 0.996	             \\
            HRNet       &	0.372&		0.490&	0.516&\textbf{	0.998}	            \\
            UNet    & 0.411  & 0.526 & 0.592 & 0.997   \\
            SOD         &	\textbf{0.445}&	\textbf{0.594}&	\textbf{0.652}&	0.997	             \\
            \hline
        \end{tabu}}
       
        \label{tab:metrics_baselines_1}
    \end{table}

     \noindent Note that despite the number of images being low, the average dimension of the images is $3480 \times 1728$ pixels.
        The width of each exported image is fixed at $1728$ pixels. 
        The height varies across the $286$ images due to it being dependent on the leg length of each survey. 
        We note that by applying the strided-cropping, as described in Section \ref{ss:post-processing}, we can obtain $2539$ images in the train set and $1722$ images in the test set of size $1728 \times 1728$ pixels. Table~\ref{tab:ship_info} provides further details of each site, including material, ship length and ship beam (width) obtained from \citet{tbnms}, as well as classification details according to our data categorization for terrain type, sonar image quality, and wreck condition.


        \begin{figure}[ht]
    \includegraphics[width=0.98\linewidth]{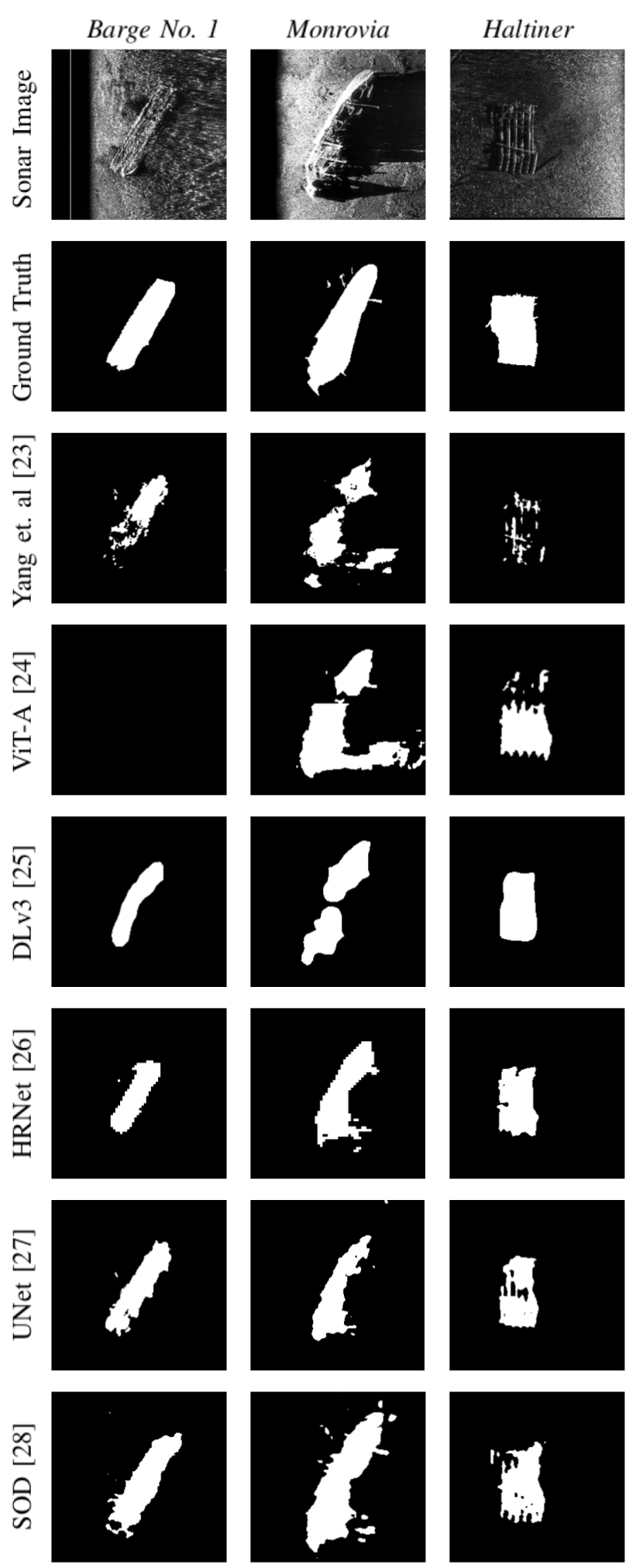}
    \caption{Shipwreck segmentation predictions from the  baselines on \textit{Barge No. 1}, \textit{Monrovia}, and \textit{Haltiner}. Zoomed in for better detail.}
     \label{fig:baseline_preds}
     \vspace{-0.5em}
    \end{figure}
        
\section{Experiments \& Results}

    \subsection{Baseline Comparison}
    We perform benchmark evaluations on a series of deep learning-based segmentation methods to evaluate the ability of state-of-the-art deep learning methods to segment shipwrecks from SSS imagery. Benchmark methods are selected to reflect the variety of deep learning-based segmentation models used in the vision community: Yang et al. \citep{baseline2}, ViT Adapter \citep{chen_vision_2023}, DeepLabV3 \citep{chen_rethinking_2017}, HRNet \citep{wang_deep_2020}, UNet \citep{ronneberger_u-net_2015}, and Salient Object Detection (SOD) InSPyReNet \citep{kim_revisiting_2022}. The model with the longest training time was SOD, which took 14 hours on a single NVIDIA A100 GPU with 80GB of memory. All baselines were trained on $512 \times 512$ images, which are downsampled from the original square cropped size of $1728 \times 1728$ described in Section \ref{ss:post-processing}. Code to replicate our cropping process and to set up a PyTorch dataset can be found in the repository linked on the project website.

    We train each baseline according to the training parameters (learning rate, epoch, batch size) suggested by their original papers. This includes the choice of encoder backbones, for which we use the pre-trained backbone associated with the highest performance reported in the original papers. Although we use pre-trained backbones, we train each model from scratch on our dataset. We use the code released from the baseline papers, except for Yang et al., which we re-implemented ourselves following the architecture discussed in \citep{baseline2}.

 \subsection{Evaluation Metrics}
   Each model is evaluated on an aggregate level with Intersection over Union (IoU) for the ``shipwreck" class ($\text{IoU}_\text{ship}$), F1 score, True Positive Rate (TPR), and True Negative Rate (TNR). All metrics are calculated in pixel-space, and are therefore measures of how well network output corresponds to expert opinions in pixel-space.
 Table~\ref{tab:metrics_baselines_1} provides results across baselines, averaged across each site. Table \ref{tab:metrics_per_site} shows $\text{IoU}_{\text{ship}}$ per site for each baseline. Segmentation predictions for all six baselines on three example sites are shown in Fig. \ref{fig:baseline_preds}. Across this comparison, the SOD model consistently outperformed or performed comparably against all other networks for the aggregate metrics. We performed the subsequent experiment with the SOD model.

    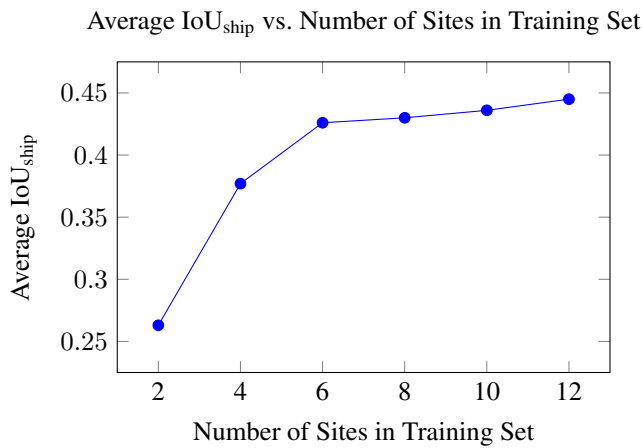
\begin{figure}[t]
    \centering
    \vspace{-0.5em}
    \begin{tikzpicture}
        \begin{axis}[
            title=Average $\text{IoU}_{\text{ship}}$ vs. Number of Sites in Training Set,
            xlabel=Number of Sites in Training Set,
            ylabel=Average $\text{IoU}_{\text{ship}}$,
            xmin=1, xmax=13,
            ymin=0.225, ymax=0.475,
            xtick={2, 4, 6, 8, 10, 12},
            xticklabels={2, 4, 6, 8, 10, 12},
            ytick={0.20, 0.25, 0.30, 0.35, 0.40, 0.45},
            height=5.7cm, width=0.95*\linewidth
                    ]
        \addplot[mark=*,blue] plot coordinates {
            (2, 0.263)
            (4, 0.377)
            (6, 0.426)
            (8, 0.430)
            (10, 0.436)
            (12, 0.445)
        };
        \end{axis}
    \end{tikzpicture}
    \caption{Results of the train set size experiment. Note the plateau in performance after six training sites. \vspace{-2mm}}
    \label{fig:size_exp}
    \end{figure}

    \subsection{Train Set Size Experiment}

    Intuitively, a larger training set should lead to better learning conditions for the network and ultimately result in a more accurate model. We conducted an experiment where we gradually increased the size of the train set back up to the full set in order to observe the relationship between train set size and model performance for our dataset. As we increase the train set size, we evaluate model performance on a frozen test set. As shown in Fig. \ref{fig:size_exp}, we observe that test performance increases with increased training dataset size and the network performance plateaus at around six sites in the train set.


\section{Conclusion \& Future Work}

This work contributes AI4Shipwrecks, an open-source dataset for comparison of state-of-the-art deep neural networks for shipwreck segmentation from SSS imagery. While recent advances in deep learning have revolutionized the field of computer vision for terrestrial robotics, adoption of similar methods across marine applications is limited due to a lack of widely available data for development and direct comparison of results. 
We establish a benchmark for semantic segmentation of shipwrecks on the AI4Shipwrecks dataset, and we provide comparison of current state-of-the-art deep neural networks for segmentation. 
The dataset and code for evaluation will be open-source to enable future research in machine learning for ocean exploration.




The AI4Shipwrecks dataset is a powerful new tool in the field of object segmentation in the wild. Field datasets are often smaller, and can help encourage discoveries in few-shot learning or simulated data generation. A promising direction for future work includes leveraging synthetic data to augment real sonar datasets for learning-based detection and segmentation tasks \citep{lee2018deep,sethuraman}. 
Additionally, AI4Shipwrecks narrows a focus on advancing network architecture to enable deep neural networks to learn from limited training data. Of notable interest is recent work that has demonstrated the potential for few shot learning for object detection from marine optical and sonar imagery \citep{ochal2020comparison}. Few shot and one shot learning approaches aim to effectively learn to represent a class of objects after seeing a few or single instance of that class, which is ideal for datasets with relatively low abundance of samples per class. We hope the AI4Shipwrecks dataset will enable future work on training from limited training data for marine applications.

\vspace{15mm}
\begin{acks}
We would like to acknowledge the lives that were lost in shipwrecks throughout Thunder Bay National Marine Sanctuary. We would also like to thank the sanctuary scientists at Thunder Bay National Marine Sanctuary for supporting field experiments and labeling efforts. Additionally, we acknowledge the effort of IJRR reviewers in providing feedback to this pre-print. The link to the official publication is: \url{https://journals.sagepub.com/doi/10.1177/02783649241266853}.
\end{acks}

\begin{dci}
The Authors declare that there is no conflict of interest.
\end{dci}

\begin{funding}
The authors disclosed receipt of the following financial support for the research, authorship, and/or publication of this article: This work is supported by the NOAA Ocean Exploration Program under award \#NA21OAR0110196.
\end{funding}

\end{document}